\documentclass{Interspeech}

\usepackage{comment}
\usepackage{multirow}
\usepackage{color}
\usepackage{cite}

\interspeechcameraready

\title{Whale: Large-Scale multilingual ASR model with w2v-BERT and E-Branchformer with large speech data}

\author[affiliation={1}]{Yosuke}{Kashiwagi}
\author[affiliation={1}]{Hayato}{Futami}
\author[affiliation={1}]{Emiru}{Tsunoo}
\author[affiliation={1}]{Satoshi}{Asakawa}

\affiliation{}{Sony Group Corporation}{Japan}
\email{\{yosuke.kashiwagi,hayato.futami,emiru.tsunoo,satoshi.asakawa\}@sony.com}

\usepackage{comment}

\begin{document}

\maketitle

\begin{abstract}
This paper reports on the development of a large-scale speech recognition model, Whale.
Similar to models such as Whisper and OWSM, Whale leverages both a large model size and a diverse, extensive dataset.
Whale’s architecture integrates w2v-BERT self-supervised model, an encoder--decoder backbone built on E-Branchformer, and a joint CTC-attention decoding strategy.
The training corpus comprises varied speech data, of not only public corpora but also in-house data, thereby enhancing the model's robustness to different speaking styles and acoustic conditions.
Through evaluations on multiple benchmarks, Whale achieved comparable performance to existing models.
In particular, it achieves a word error rate of 2.4\% on the Librispeech test-clean set and a character error rate of 3.4\% on the CSJ eval3 set, outperforming Whisper large-v3 and OWSM v3.1.
%In addition, we investigated the case on out-of-domain data and effectiveness of zero-shot language adaptation using self-conditioned CTC.
%These results highlight Whale’s effectiveness in large-scale, multilingual speech recognition tasks.
%On the other hand, we also examined zero-shot language adaptation using self-conditioned CTC. However the effect was not confirmed in this large-scale model.
\end{abstract}

\section{Introduction}

Recent breakthroughs in multilingual ASR owe a great deal to the open-source community and researchers who have worked tirelessly to collect, curate, and share large-scale multilingual speech datasets.
Projects such as CommonVoice, MuST-C, mTEDx, MLS, YODAS, FLERUS and others~\cite{commonvoice_ardila2020common,mustc_di2019must,mtedx_elizabeth2021multilingual,mls_pratap2020mls, yodas_li2023yodas,fleurs_conneau2023fleurs,jtubespeech_takamichi2021jtubespeech,vctk_yamaghishi2017vctk} have enabled model developers to train and evaluate systems on diverse languages and dialects, thus fostering inclusivity and advancing ASR for low-resource languages.
Their collective contributions have fundamentally reshaped the ASR research landscape, making it possible for anyone to experiment with large-scale data and accelerate innovations in multilingual recognition.

Building on these important contributions, the past a few years have also witnessed the emergence of several multilingual speech recognition models that have pushed the state-of-the-art in performance~\cite{multiasr_toshniwal2018multilingual,multiasr_li2019bytes,multiasr_seki2018end,multiasr_pratap2024scaling,multiasr_puvvada2024less}.
For example, Whisper~\cite{whisper_radford2023robust} represents one of the most notable achievements in the field.
It leverages an end-to-end framework trained on a diverse set of multilingual data, yielding high accuracy in both clean and noisy conditions.
However, despite its strong performance, many details about the training data, procedure and hyperparameter tuning remain proprietary, which prevent the community from fully reproducing and build upon its results.

In contrast, other models such as OWSM and OWSM CTC~\cite{owsm_peng2024owsm,owsmv1_peng2023reproducing,owsmctc_peng2024owsmctc} have been developed with a more open approach.
OWSM provides valuable insights into model design and training on a moderate scale.
Nevertheless, the training data and computational resources utilized for OWSM are relatively limited compared to those employed by Whisper, and the experience derived from extensive real-world data has not been as thoroughly explored.
These gaps highlight a broader challenge in the ASR community: how to effectively harness large-scale data and computational resources while maintaining transparency and reproducibility in research.

Against this backdrop, our work introduces Whale, a large-scale speech recognition model that aims to bridge the gap between proprietary high-performance systems and openly accessible models.
The design of Whale is based on three architectural innovations.
First, Whale utilizes a self-supervised learning (SSL) framework, specifically employing the w2v-BERT model with 24 Conformer layers~\cite{w2vbert_chung2021w2v}.
W2v-BERT captures intricate patterns in the raw speech signal, providing high-quality features that serve as input to the subsequent modules.
Second, attention-based encoder--decoder model is adopted.
An encoder module, built with 24 layers of E-Branchformer blocks~\cite{ebranchformer_kim2023branchformer}, is responsible for encoding these features into a latent representation.
Complementing the encoder, a six layer Transformer decoder refines the output, generating accurate transcription sequences.
Third, to further enhance performance, CTC branch is integrated into the network, facilitating a joint-decoding approach during inference that combines the strengths of both CTC and attention-based decoding strategies~\cite{jointdecoding_kim2017joint,jointdecoding_hori2017joint}.

The total amount of our training data is 250k, covering 144 languages.
The training data combines large, publicly available multilingual corpora with substantial in-house recordings.
This stance is similar to that of Canary~\cite{multiasr_puvvada2024less}, but we stress that the total amount of Japanese data is notably higher in our collection.
Because public data is biased towards English, we collected Japanese data as one of our targets.
In other words, almost all of the data other than Japanese is publicly available data.
Therefore, we can expect the effect of adding in-house data to be clear by comparing Japanese and other languages.
%By merging open-source and proprietary datasets, we expect broad coverage of accents, speaking styles, and acoustic scenarios, helping the model achieve robust performance.

Furthermore, the training strategy adopted for Whale addresses the training stability in large-scale ASR model development.
Due to the inherent instability in training multilingual models with massive parameter counts, we employ a curriculum learning approach~\cite{curriculum_bengio2009curriculum}.
Initially, the model is trained on simpler, single-language tasks using a reduced network capacity.
Gradually, the training regimen is scaled up to incorporate more languages and complex acoustic scenarios.
This staged learning process not only improves convergence but also enhances the overall robustness of the model.
Additionally, after the initial phase, previously frozen parameters in the w2v-BERT module are updated to better align with the ASR objective.
%, further refining the model's performance.

We evaluate Whale on multiple standard benchmarks, including Librispeech, CSJ, CommonVoice, and FLEURS corpus\cite{librispeech_panayotov2015librispeech,csj_maekawa2003corpus,commonvoice_ardila2020common,fleurs_conneau2023fleurs}.
Experimental results demonstrate that Whale achieved comparable performance compared with state-of-the-art systems, such as Whisper and OWSM, in terms of word or character error rates across various languages.

\section{Model architecture}

Our model assumes a single-channel input speech, and if the sampling rate differs, it is resampled to 16kHz.
Next, acoustic features are extracted via SSL, w2v-BERT~\cite{w2vbert_chung2021w2v}.
At this point, the frame rate is 100 frames/sec.
The features are then input into an encoder based on E-Branchformer~\cite{ebranchformer_kim2023branchformer}.
In the encoder, convolution layers first reduce the frame rate by half through convolution operations, followed by the application of 24 layers of E-Branchformer blocks.
Thereby, the frame rate of the encoder output is 50 frames/sec.
In the 8th and 16th blocks, self-conditioned CTC~\cite{scctc_komatsu2022better} is used, which is employed not only for increasing performance but also for zero-shot language adaptation during inference, as described later.
The output from the encoder is then passed to the decoder.
The decoder consists of six layers of transformer blocks.
Additionally, the encoder's output is also input to a CTC branch, which is used for joint-decoding~\cite{jointdecoding_kim2017joint,jointdecoding_hori2017joint} along with the decoder output during inference.
The total number of parameters is 1.87B including 0.58B w2v-BERT parameters.

\subsection{w2v-BERT}

w2v-BERT is a self-supervised learning model designed for extracting robust speech representations that capture both local acoustic patterns and long-range contextual dependencies~\cite{w2vbert_chung2021w2v}.
In our system, the w2v-BERT module is implemented with 24 layers, each consisting of a Conformer block.
The model is trained using a masked prediction strategy inspired by the BERT~\cite{bert_devlin2018bert} framework in natural language processing.
During pre-training, segments of the input speech are randomly masked, and the model is tasked with predicting the missing parts based on the surrounding unmasked context.
This approach encourages the model to learn deep, contextualized representations of the speech data.
In addition to the masked prediction loss, a contrastive loss is employed to further distinguish true speech representations from negative samples.
This dual-objective training ensures that the learned features are rich in information enough for a variety of downstream tasks.

By incorporating pre-trained w2v-BERT\footnote{https://huggingface.co/facebook/w2v-bert-2.0} into our framework, we harness its powerful feature extraction capabilities, which are critical for capturing the intricate patterns in speech data.
Other multilingual models are also available, but they have issues such as small model size and low language coverage~\cite{mhubert_boito2024mhubert,wavlm_chen2022wavlm}.
These high-quality representations serve as the foundation for subsequent processing in our model, enabling Whale to achieve superior performance in large-scale speech recognition tasks.
Furthermore, we are not just using it as a feature extraction, but we are also enhancing its effectiveness by jointly finetuning it with the objective function of ASR.

\subsection{Encoder--Decoder model}

Our encoder--decoder framework is designed to efficiently capture both local acoustic nuances and long-range dependencies in speech signals.
The system is composed of two main components: a robust encoder built with E-Branchformer blocks, and a Transformer-based decoder.

\subsubsection{Encoder: E-Branchformer}

The encoder leverages 24 layers of E-Branchformer blocks~\cite{ebranchformer_kim2023branchformer}.
E-Branchformer is specifically designed for speech recognition tasks by expanding the capacity of traditional Transformer architectures.
Its key innovation lies in the introduction of multiple parallel branches within each block, enabling the model to capture multi-scale features and diverse temporal dynamics.
Each block integrates:

\begin{itemize}
    \item \textbf{Multi-Scale Feature Extraction:} Parallel branches process the input representation at different resolutions or with varied receptive fields, effectively capturing both fine-grained local patterns and broader contextual cues.
    \item \textbf{Convolutional Sub-Layers:} Convolution operations are employed to reduce the frame rate and to emphasize local temporal dependencies, providing complementary information to the self-attention mechanisms.
    \item \textbf{Self-Attention Mechanisms:} The self-attention layers facilitate global context modeling, ensuring that the dependencies across distant time steps are well captured.
    \item \textbf{Fusion Strategy:} The outputs from the multiple branches are fused using learned weighting mechanisms, allowing the network to dynamically emphasize the most relevant features.
\end{itemize}

%In our implementation, the initial convolution layers in the encoder reduce the frame rate by half, after which the 24-layer E-Branchformer progressively refines the representations.
%Notably, in the 8th and 16th layers, self-conditioned CTC~\cite{scctc_komatsu2022better} is incorporated to enable zero-shot language adaptation, as detailed below.

\subsubsection{Self-Conditioned CTC for Language Adaptation}

To enhance multilingual performance and enable zero-shot language adaptation, we integrate self-conditioned CTC~\cite{scctc_komatsu2022better} into the encoder in the 8th and 16th layers.
Self-conditioned CTC conditions the model's predictions on its previous outputs to iteratively refine the intermediate representations.
This mechanism works as follows:

\begin{enumerate}
    \item \textbf{Intermediate Predictions:} At designated layers (specifically the 8th and 16th E-Branchformer blocks), the model generates preliminary predictions via CTC branches.
          These predictions serve as a soft alignment signal.
    \item \textbf{Feedback Mechanism:} The predicted outputs are fed back into the network to condition the subsequent layers.
          This self-conditioning adjusts the latent representations by effectively providing language-specific cues, which helps in adapting to the phonetic and linguistic characteristics of the target language without the need for explicit fine-tuning.
    \item \textbf{Dynamic Language Adaptation:} By modulating the probability distributions based on the target language ID during inference, the self-conditioned CTC mechanism allows the model to implicitly re-calibrate its outputs, thereby facilitating zero-shot adaptation across multiple languages~\cite{encoderprompting_kashiwagi2024rapid}.
\end{enumerate}

\begin{table}[t]
\caption{Training data hours for top-5 languages in each resource rank. High is for languages
with more than 100 hours of data, middle is for 20-100 hours, low is for less than 20 hours. The blanks are unknown. In the case of Pashto, Whisper only uses translation data.}
\label{table:trainingdata}
\centering
\begin{tabular}{clcccc}
\hline
 & \textbf{Language} & \textbf{LID} & Whale & Whisper & OWSM \\
\hline
\multirow{5}{*}{High} 
 & English    & en & 81k & 438k & 73k \\
 & Japanese   & ja & 30k & 7k & 19k \\
 & Mandarin   & zh & 11k & 23k & 16k \\
 & German     & de & 2k  & 13k & 3.7k \\
 & French     & fr & 2k  & 9k & 2.5k \\
\hline
\multirow{5}{*}{Middle} 
 & Czech        & cs  & 84 & 192 & --\\
 & Romanian     & ro  & 75 & 356 & --\\
 & Swahilli     & sw  & 75 & 5.4 & --\\
 & Thai         & th  & 71 & 226 & --\\
 & Ukranian     & uk  & 45 & 697 & --\\
\hline
\multirow{5}{*}{Low} 
 & Korean     & ko  & 19  & 8k & --\\
 & Welsh      & cy  & 19  & 73 & --\\
 & Telugu     & te  & 19  & 4.3 & --\\
 & Cantonese  & yue & 18  & 23k & --\\
 & Pashto     & ps  & 17  & 0* & --\\
\hline
\end{tabular}
\end{table}

\begin{comment}
\begin{figure}[tbp]
\begin{center}
\includegraphics[width=\linewidth]{fig/acc.png}
\caption{Training acc curve for stages 6--7.}
\end{center}
\end{figure}
\end{comment}

\subsubsection{Decoder: Transformer}

Complementing the encoder, the decoder is implemented as a six layer Transformer.
The decoder refines the encoder’s output and generates the final transcription sequence.
A key aspect of our design is the joint training of the CTC branch and the Transformer decoder:

\begin{itemize}
    \item \textbf{CTC Branch Integration:} The encoder's output is passed through a linear projection layer that connects to a CTC objective.
          During training, the loss from this branch is combined with the decoder loss using a weighted sum, where we assign a weight of 0.3 to the CTC loss and 0.7 to the decoder loss.
    \item \textbf{Joint-Decoding Strategy:} During inference, beam search is conducted by integrating scores from both the CTC branch and the decoder.
          The CTC branch is used as a prefix search to guide the beam search, and the final score for each candidate is computed by summing the weighted contributions (0.3 for CTC, 0.7 for the decoder).
\end{itemize}

%In summary, our Encoder-Decoder model builds upon the strengths of E-Branchformer for multi-scale feature extraction and leverages self-conditioned CTC for effective zero-shot language adaptation.
%This combination, along with a robust Transformer-based decoder and a joint-decoding strategy, results in a model that is highly effective for large-scale, multilingual speech recognition.

\section{Training}

\subsection{Training environments}

The training of the Whale model was conducted on an internal server infrastructure.
Each node in our cluster is equipped with eight NVIDIA H100 GPUs, and we were able to leverage a total of 16 nodes simultaneously.
The entire training process spanned approximately six weeks.
Our implementation and training pipeline are built upon ESPNet~\cite{watanabe18_interspeech}, which is based on PyTorch.
ESPNet has the advantage of being able to easily conduct a training by connecting multiple modules, making it ideal for training large-scale multilingual speech recognition models.
This environment provided the necessary computational resources and scalability to efficiently handle the large-scale data and complex model architectures employed in our work.

\subsection{Training data}

The training data for Whale is divided into three ranks (High is for languages with more than 100 hours of data, Middle is for 20-100 hours, Low is for less than 20 hours).
%, based on how many hours of data each language has:
%\begin{itemize}
%    \item \textbf{High:} more than 100 hours
%    \item \textbf{Middle:} 20--100 hours
%    \item \textbf{Low:} less than 20 hours
%\end{itemize}
Table~\ref{table:trainingdata} shows examples of the top five languages in each rank.
Overall, our data totals 250k hours and covers 144 languages.

The sources of this data fall into three categories:
\begin{enumerate}
    \item \textbf{Open data:}
          We incorporate publicly available corpora including Common Voice~\cite{commonvoice_ardila2020common},
          Librispeech~\cite{librispeech_panayotov2015librispeech}, MLS~\cite{mls_pratap2020mls}, VoxPopuli~\cite{voxpopuli_wang-etal-2021-voxpopuli}, WenetSpeech~\cite{wenet_zhang2022wenetspeech}, People’s Speech~\cite{peoples_galvez2021people}, FLEURS~\cite{fleurs_conneau2023fleurs},
          JTubeSpeech~\cite{jtubespeech_takamichi2021jtubespeech}, YODAS~\cite{yodas_li2023yodas}, 
          and NPSC~\cite{npsc_solberg2022norwegian}, among others.
    \item \textbf{Purchased datasets:}
          Commercial datasets obtained under standard licenses including WSJ\footnote{https://catalog.ldc.upenn.edu/LDC93S6A}, CSJ~\cite{csj_maekawa2003corpus}, King-ASR~\footnote{https://dataoceanai.com/}, ELRA Speecon~\footnote{https://catalogue.elra.info/en-us/}, among others.
    \item \textbf{Independently collected and curated in-house data:}
          Proprietary recordings, which are
          almost entirely in Japanese.
\end{enumerate}

All data sources are acquired under commercially available licenses, 
ensuring that the resulting model can be deployed in real-world applications 
without additional licensing concerns.
Despite this extensive corpus, the distribution of data remains unbalanced 
across languages, reflecting the broader challenge of multilingual ASR research.

%Based on our empirical studies, we determined that a minimum of 20 hours of speech data per language is essential to achieve stable and reliable recognition performance.
%Consequently, only 58 out of the 144 languages in our corpus meet this criterion, which has implications for both the performance and adaptability of the model across different languages.

\begin{table}[t!]
\caption{Comparison on Librispeech data (WER \%). The numbers represent the dev/test sets respectively.}
  \label{table:results_librispeech}
  \centering
  \begin{tabular}{lc|cc}
    \hline
                     & size & clean & other \\ \hline
    Whisper large-v3 & 1.55B & 3.5/2.5 & \bf{4.2/4.3}\\
    OWSM v3.1        & 1.02B & 2.3/2.4 & 4.9/5.0 \\
    OWSM CTC v3.1    & 1.01B & \bf{2.2/2.4} & 5.1/5.1 \\
    Whale stage6     & 1.87B & 2.5/2.5 & 6.0/5.8  \\
    Whale stage7     & 1.87B & \bf{2.2/2.4} & 4.5/4.5  \\
    \hline
  \end{tabular}
\end{table}

\begin{table}[t!]
\caption{Comparison on CSJ data (CER \%). }
  \label{table:results_csj}
  \centering
  \begin{tabular}{lc|ccc}
    \hline
                     & size & eval1 & eval2 & eval3 \\ \hline
    Whisper large-v3 & 1.55B & 18.4  & 17.5  & 16.5  \\
    OWSM v3.1        & 1.02B & 30.5  & 26.7  & 21.9  \\
    OWSM CTC v3.1    & 1.01B & 33.9  & 29.8  & 26.1  \\
    Whale            & 1.87B & \bf{\phantom{0}5.1}   & \bf{\phantom{0}4.2}   & \bf{\phantom{0}3.4}   \\
    \hline
  \end{tabular}
\end{table}

\begin{table}[t!]
\caption{Comparison on Commonvoice data. The * mark indicates CER, and the rest are WER.}
  \label{table:results_commonvoice}
  \footnotesize
  \centering
  \begin{tabular}{lc|ccccc}
    \hline
                      & size & en  & ja*  & zh*  & de   & fr \\ \hline
    Whisper large-v3 & 1.55B & \bf{\phantom{0}8.4}  & 12.2 & 12.8 & \bf{\phantom{0}6.4}  & 11.0 \\
    OWSM v3.1        & 1.02B & 11.2 & 17.5 & 20.2   & 11.0 & 13.9   \\
    OWSM CTC v3.1    & 1.01B & 10.6 & 12.3 & 17.4 & 11.9 & 14.5 \\
    Whale            & 1.87B & \phantom{0}8.9  & \bf{11.7} & \bf{11.0} & \bf{\phantom{0}6.4}  & \bf{\phantom{0}9.0}  \\
    \hline
  \end{tabular}
\end{table}

\begin{table*}[t!]
\caption{Comparison on FLEURS data. The * mark indicates CER, and the rest are WER. The blanks indicate unsupported languages.}
  \label{table:results_fleurs}
  \footnotesize
  \centering
  \begin{tabular}{lc|ccccc|ccccc|ccccc}
    \hline
                      & size & en & ja* & zh* & de & fr & cs & ro & sw & th & uk & ko & cy & te & yue* & ps \\ \hline
    Whisper large-v3 & 1.55B & \bf{4.0} & \bf{4.5} & \bf{7.8} & \bf{5.8} & \bf{5.3} & \bf{11.0} & \bf{8.2} & 34.5 & \bf{30.0} & \bf{11.5} & \bf{14.1} & \bf{27.9} & 38.0 & \bf{9.8} & 89.0 \\
    OWSM v3.1        & 1.02B & 7.2 & 9.1 & 32.1 & 12.9 & 16.3 & -- & 41.7 & 44.0 & 83.2 & 52.2 & 57.1 & 45.2 & 61.2 & 48.3 & 84.9 \\
    OWSM CTC v3.1    & 1.01B & 8.2 & 7.6 & 14.6 & 15.3 & 17.1 & -- & 44.8 & 102.1 & 69.8 & 47.4 & 50.7 & 47.3 & 45.6 & 41.6 & 79.4 \\
    Whale            & 1.87B & 6.2 & 4.9 & 10.6 & 9.5 & 9.7 & 14.9 & 23.5 & 18.0 & 37.9 & 16.3& 70.9 & 41.2 & \bf{15.1} & 26.1 & \bf{44.1} \\
    + adaptation     & 1.87B & 6.4 & 5.4 & 11.8 & 10.0 & 9.8 & 14.2 & 38.6 & \bf{17.8} & 38.9 & 16.4 & 74.9 & 41.6 & 16.3 & 28.7 & 47.3 \\
    \hline
  \end{tabular}
\end{table*}

\subsection{Curriculum learning}

To improve training stability, especially in the context of multilingual speech recognition with a large network, curriculum learning is employed \cite{curriculum_bengio2009curriculum}.
In our approach, training is structured into seven stages, gradually increasing the model complexity and the diversity of the training data.
This staged training allows the model to first learn robust representations on simpler tasks before being exposed to more challenging multilingual data.
The curriculum is organized as follows:

\begin{itemize}
    \item \textbf{Stage 1:} Train a small (8-layer) encoder model using only CommonVoice English data for 1 day.
    \item \textbf{Stage 2:} Train a medium-sized (16-layer) encoder model using CommonVoice English data for 1 day.
    \item \textbf{Stage 3:} Train a large (24-layer) encoder model using CommonVoice English data for 1 day.
    \item \textbf{Stage 4:} Train the large encoder model with subsampled data from all available English data for 1 day.
    \item \textbf{Stage 5:} Train the large encoder model using CommonVoice multilingual data for 3 days.
    \item \textbf{Stage 6:} Train the large encoder model with all available data across languages for 2 weeks.
    \item \textbf{Stage 7:} Continue training the large encoder model with SSL updating (i.e., updating previously frozen parameters) for 3 weeks.
\end{itemize}

\section{Evaluation}

The primary objective of our evaluation is to rigorously assess the performance and robustness of the Whale model across diverse datasets and acoustic conditions.
Our experiments compare Whale against state-of-the-art systems such as Whisper~\cite{whisper_radford2023robust}, OWSM~\cite{owsm_peng2024owsm}, and OWSM CTC~\cite{owsmctc_peng2024owsmctc}.
%Additionally we focused on whether the in-house Japanese data contributes to performance in large-scale, multilingual settings.
Before calculating the scores, we applied the whisper-normalizer\footnote{https://github.com/openai/whisper} to both the references and the hypotheses.
%We also provide initial observations on zero-shot adaptation potential to handle a variety of languages and investigate the feasibility in our large model.
%While the model employs both a joint-decoding strategy and a curriculum learning approach, we do not perform direct ablations on these methods here; instead, we present overall performance outcomes with the model’s full configuration.

\subsection{Librispeech}

Table~\ref{table:results_librispeech} shows the Word Error Rate (WER) comparison on the English ASR corpora, Librispeech dataset \cite{librispeech_panayotov2015librispeech}.
We evaluated on both the clean and other sets.
Whisper large-v3 demonstrated strong performance with WERs of 3.5\%/2.5\% on dev-clean/test-clean and 4.2\%/4.3\% on dev-other/test-other, respectively.
OWSM v3.1 and OWSM CTC v3.1 achieved slightly better results on the clean sets, but their performance on the other sets was around 5.0\%.

Our Whale model is reported at two training stages: stage6 and stage7.
After the large-scale curriculum training (stage6), Whale obtains a WER of 2.5\% for dev-clean and test-clean, comparable to Whisper and OWSM.
However, its performance on dev-other/test-other (6.0\%/5.8\%) indicates a room for improvement.
Upon further SSL parameter updating (stage7), Whale reaches 2.2\%/2.4\% for dev-clean/test-clean and 4.5\%/4.5\% on dev-other/test-other, surpassing OWSM and closing the gap against Whisper on the other subsets.
Therefore, we can see that updating SSL is extremely important even for such a huge model.

\subsection{CSJ}

To evaluate performance on Japanese spontaneous speech, we use the CSJ dataset~\cite{csj_maekawa2003corpus}.
As shown in Table~\ref{table:results_csj}, we compared character error rates (CER) on the standard eval1, eval2, and eval3 sets.
Whisper large-v3 achieves CERs between 16.5\% and 18.4\%, while OWSM v3.1 and OWSM CTC v3.1 produce CERs in the 20--30\% range, indicating difficulty in handling spontaneous Japanese speech.
It is also important that the CSJ is the audio of a simulated lecture.
There are many fillers and hesitations, and for example, Whisper tends to delete them.
Whale demonstrates significantly lower CERs (5.1\%, 4.2\%, and 3.4\% on eval1, eval2, and eval3, respectively).
These results show the effect of collecting Japanese-language data more intensively than Whisper and OWSM.

\subsection{Commonvoice}

Next, we examine recognition performance on the Commonvoice dataset~\cite{commonvoice_ardila2020common}.
Table~\ref{table:results_commonvoice} reports WER for English, German, and French, while Japanese and Chinese metrics are shown as CER (marked with an asterisk).
Whisper large-v3 achieves strong results for English (8.4\% WER), but yields higher error rates for French (11.0\% WER) and Chinese (12.8\% CER).
OWSM v3.1 performs worse overall, particularly on Chinese (20.2\% CER) and French (13.9\% WER).
Its CTC variant slightly improves the error rate for Chinese to 17.4\% but remains behind Whisper.
In comparison, Whale maintains competitive WERs across all evaluated languages, especially for Japanese (11.7\% CER) and Chinese (11.0\% CER).
Notably, Whale equals Whisper on German (6.4\% WER) and delivers a lower French WER of 9.0\%.

\subsection{FLEURS}

Finally, Table~\ref{table:results_fleurs} presents WER/CER on a selection of languages described in Table \ref{table:trainingdata}.
%Here, the * mark indicates CER, while other columns represent WER.
Whisper large-v3 generally performed well for many languages.
This was a great performance backed up by the amount of training data.
For this reason, Whale achieved better performance in some languages with small amounts of data (e.g. Swahilli and Telugu).
OWSM v3.1 and OWSM CTC v3.1 show even higher error rates on many languages.
Unlike commonvoice, a significant difference was seen between Whale and Whisper on FLEURS data.
Whisper performed better, even in Japanese that was particularly collected in large scale.
Because most of our in-house data was read speech, we considered this as a problem of robustness against the data out of domain.
%We considered this to be a problem of robustness to data outside the domain.
%Actually, much of our in-house data was read speech.

In addition, we also performed the zero-shot language adaptation described in Sec. 2.2.2.
However, it did not have any effect on most languages.
This is thought to be due not only to the effect of domain mismatch, but also to w2v-BERT SSL model.
It is thought that there is no room for adaptation because a huge SSL network exist before the encoder.

\section{Conclusion}

In this paper, we introduced Whale, a large-scale speech recognition model that leverages a robust encoder--decoder architecture, self-conditioned CTC, and a carefully designed training strategy including curriculum learning and SSL updating.
Through extensive experiments, we demonstrated that Whale achieves highly competitive performance on four benchmarks.
On the other hand, we also confirmed that Whisper shows good results in FLERUS data, suggesting the need for expanding domain range.
We also confirmed that zero-shot language adaptation was not effective in such a huge network.

Future work will explore further optimizations in model compression and inference speed, as well as expanded language support.
Additionally, we plan to investigate such as data augmentation techniques and domain adaptation strategies to maintain performance in real-world, noisy environments.
We hope that making the details of Whale’s design and training methodology transparent will encourage deeper engagement from the research community and foster continued progress in large-scale, multilingual speech recognition.

\newpage
\bibliographystyle{IEEEtran}
\bibliography{mybib}

\begin{thebibliography}{10}
\providecommand{\url}[1]{#1}
\csname url@samestyle\endcsname
\providecommand{\newblock}{\relax}
\providecommand{\bibinfo}[2]{#2}
\providecommand{\BIBentrySTDinterwordspacing}{\spaceskip=0pt\relax}
\providecommand{\BIBentryALTinterwordstretchfactor}{4}
\providecommand{\BIBentryALTinterwordspacing}{\spaceskip=\fontdimen2\font plus
\BIBentryALTinterwordstretchfactor\fontdimen3\font minus \fontdimen4\font\relax}
\providecommand{\BIBforeignlanguage}[2]{{%
\expandafter\ifx\csname l@#1\endcsname\relax
\typeout{** WARNING: IEEEtran.bst: No hyphenation pattern has been}%
\typeout{** loaded for the language `#1'. Using the pattern for}%
\typeout{** the default language instead.}%
\else
\language=\csname l@#1\endcsname
\fi
#2}}
\providecommand{\BIBdecl}{\relax}
\BIBdecl

\bibitem{commonvoice_ardila2020common}
R.~Ardila, M.~Branson, K.~Davis \emph{et~al.}, ``{Common Voice: A Massively-Multilingual Speech Corpus},'' in \emph{Proceedings of the Twelfth Language Resources and Evaluation Conference}, 2020, pp. 4218--4222.

\bibitem{mustc_di2019must}
M.~A. Di~Gangi, R.~Cattoni, L.~Bentivogli, M.~Negri, and M.~Turchi, ``{MUST-C: a multilingual speech translation corpus},'' in \emph{Proceedings of the 2019 Conference of the North American Chapter of the Association for Computational Linguistics: Human Language Technologies, Volume 1 (Long and Short Papers)}.\hskip 1em plus 0.5em minus 0.4em\relax Association for Computational Linguistics, 2019, pp. 2012--2017.

\bibitem{mtedx_elizabeth2021multilingual}
S.~Elizabeth, W.~Matthew, B.~Jacob \emph{et~al.}, ``{The Multilingual TEDx Corpus for Speech Recognition and Translation},'' in \emph{Proceedings of Interspeech 2021}, 2021, pp. 3655--3659.

\bibitem{mls_pratap2020mls}
V.~Pratap, Q.~Xu, A.~Sriram, G.~Synnaeve, and R.~Collobert, ``{MLS: A Large-Scale Multilingual Dataset for Speech Research},'' 2020.

\bibitem{yodas_li2023yodas}
X.~Li, S.~Takamichi, T.~Saeki \emph{et~al.}, ``{YODAS: YouTube-oriented dataset for audio and speech},'' in \emph{2023 IEEE Automatic Speech Recognition and Understanding Workshop (ASRU)}.\hskip 1em plus 0.5em minus 0.4em\relax IEEE, 2023, pp. 1--8.

\bibitem{fleurs_conneau2023fleurs}
A.~Conneau, M.~Ma, S.~Khanuja \emph{et~al.}, ``{FLEURS: Few-shot learning evaluation of universal representations of speech},'' in \emph{2022 IEEE Spoken Language Technology Workshop (SLT)}.\hskip 1em plus 0.5em minus 0.4em\relax IEEE, 2023, pp. 798--805.

\bibitem{jtubespeech_takamichi2021jtubespeech}
S.~Takamichi, L.~K{\"u}rzinger, T.~Saeki, S.~Shiota, and S.~Watanabe, ``{JTubeSpeech: corpus of Japanese speech collected from YouTube for speech recognition and speaker verification},'' \emph{arXiv preprint arXiv:2112.09323}, 2021.

\bibitem{vctk_yamaghishi2017vctk}
\BIBentryALTinterwordspacing
C.~Veaux, J.~Yamagishi, and K.~MacDonald, ``{CSTR VCTK Corpus: English Multi-speaker Corpus for CSTR Voice Cloning Toolkit},'' 2017. [Online]. Available: \url{https://doi.org/10.7488/ds/1994}
\BIBentrySTDinterwordspacing

\bibitem{multiasr_toshniwal2018multilingual}
S.~Toshniwal, T.~N. Sainath, R.~J. Weiss \emph{et~al.}, ``{Multilingual speech recognition with a single end-to-end model},'' in \emph{2018 IEEE international conference on acoustics, speech and signal processing (ICASSP)}.\hskip 1em plus 0.5em minus 0.4em\relax IEEE, 2018, pp. 4904--4908.

\bibitem{multiasr_li2019bytes}
B.~Li, Y.~Zhang, T.~Sainath, Y.~Wu, and W.~Chan, ``{Bytes are all you need: End-to-end multilingual speech recognition and synthesis with bytes},'' in \emph{ICASSP 2019-2019 IEEE International Conference on Acoustics, Speech and Signal Processing (ICASSP)}.\hskip 1em plus 0.5em minus 0.4em\relax IEEE, 2019, pp. 5621--5625.

\bibitem{multiasr_seki2018end}
H.~Seki, S.~Watanabe, T.~Hori, J.~Le~Roux, and J.~R. Hershey, ``{An end-to-end language-tracking speech recognizer for mixed-language speech},'' in \emph{2018 IEEE international conference on acoustics, speech and signal processing (ICASSP)}.\hskip 1em plus 0.5em minus 0.4em\relax IEEE, 2018, pp. 4919--4923.

\bibitem{multiasr_pratap2024scaling}
V.~Pratap, A.~Tjandra, B.~Shi \emph{et~al.}, ``{Scaling speech technology to 1,000+ languages},'' \emph{Journal of Machine Learning Research}, vol.~25, no.~97, pp. 1--52, 2024.

\bibitem{multiasr_puvvada2024less}
K.~C. Puvvada, P.~{\.Z}elasko, H.~Huang \emph{et~al.}, ``{Less is More: Accurate Speech Recognition \& Translation without Web-Scale Data},'' in \emph{Proc. Interspeech 2024}, 2024, pp. 3964--3968.

\bibitem{whisper_radford2023robust}
A.~Radford, J.~W. Kim, T.~Xu \emph{et~al.}, ``{Robust speech recognition via large-scale weak supervision},'' in \emph{International conference on machine learning}.\hskip 1em plus 0.5em minus 0.4em\relax PMLR, 2023, pp. 28\,492--28\,518.

\bibitem{owsm_peng2024owsm}
Y.~Peng, J.~Tian, W.~Chen \emph{et~al.}, ``{OWSM v3.1: Better and faster open whisper-style speech models based on e-branchformer},'' \emph{arXiv preprint arXiv:2401.16658}, 2024.

\bibitem{owsmv1_peng2023reproducing}
Y.~Peng, J.~Tian, B.~Yan \emph{et~al.}, ``Reproducing whisper-style training using an open-source toolkit and publicly available data,'' in \emph{2023 IEEE Automatic Speech Recognition and Understanding Workshop (ASRU)}.\hskip 1em plus 0.5em minus 0.4em\relax IEEE, 2023, pp. 1--8.

\bibitem{owsmctc_peng2024owsmctc}
Y.~Peng, Y.~Sudo, M.~Shakeel, and S.~Watanabe, ``{OWSM-CTC: An Open Encoder-Only Speech Foundation Model for Speech Recognition, Translation, and Language Identification},'' \emph{arXiv preprint arXiv:2402.12654}, 2024.

\bibitem{w2vbert_chung2021w2v}
Y.-A. Chung, Y.~Zhang, W.~Han \emph{et~al.}, ``{W2v-BERT: Combining contrastive learning and masked language modeling for self-supervised speech pre-training},'' in \emph{2021 IEEE Automatic Speech Recognition and Understanding Workshop (ASRU)}.\hskip 1em plus 0.5em minus 0.4em\relax IEEE, 2021, pp. 244--250.

\bibitem{ebranchformer_kim2023branchformer}
K.~Kim, F.~Wu, Y.~Peng \emph{et~al.}, ``{E-Branchformer: Branchformer with enhanced merging for speech recognition},'' in \emph{2022 IEEE Spoken Language Technology Workshop (SLT)}.\hskip 1em plus 0.5em minus 0.4em\relax IEEE, 2023, pp. 84--91.

\bibitem{jointdecoding_kim2017joint}
S.~Kim, T.~Hori, and S.~Watanabe, ``{Joint CTC-attention based end-to-end speech recognition using multi-task learning},'' in \emph{2017 IEEE international conference on acoustics, speech and signal processing (ICASSP)}.\hskip 1em plus 0.5em minus 0.4em\relax IEEE, 2017, pp. 4835--4839.

\bibitem{jointdecoding_hori2017joint}
T.~Hori, S.~Watanabe, and J.~R. Hershey, ``{Joint CTC/attention decoding for end-to-end speech recognition},'' in \emph{Proceedings of the 55th Annual Meeting of the Association for Computational Linguistics (Volume 1: Long Papers)}, 2017, pp. 518--529.

\bibitem{curriculum_bengio2009curriculum}
Y.~Bengio, J.~Louradour, R.~Collobert, and J.~Weston, ``Curriculum learning,'' in \emph{Proceedings of the 26th annual international conference on machine learning}, 2009, pp. 41--48.

\bibitem{librispeech_panayotov2015librispeech}
V.~Panayotov, G.~Chen, D.~Povey, and S.~Khudanpur, ``{Librispeech: an asr corpus based on public domain audio books},'' in \emph{2015 IEEE international conference on acoustics, speech and signal processing (ICASSP)}.\hskip 1em plus 0.5em minus 0.4em\relax IEEE, 2015, pp. 5206--5210.

\bibitem{csj_maekawa2003corpus}
K.~Maekawa, ``{Corpus of Spontaneous Japanese: Its design and evaluation},'' in \emph{ISCA \& IEEE Workshop on Spontaneous Speech Processing and Recognition}, 2003.

\bibitem{scctc_komatsu2022better}
T.~Komatsu, Y.~Fujita, J.~Lee \emph{et~al.}, ``{Better Intermediates Improve CTC Inference},'' in \emph{Proceedings of the Annual Conference of the International Speech Communication Association, INTERSPEECH}, vol. 2022, 2022, pp. 4965--4969.

\bibitem{bert_devlin2018bert}
J.~Devlin, ``{BERT: Pre-training of deep bidirectional transformers for language understanding},'' \emph{arXiv preprint arXiv:1810.04805}, 2018.

\bibitem{mhubert_boito2024mhubert}
M.~Z. Boito, V.~Iyer, N.~Lagos, L.~Besacier, and I.~Calapodescu, ``{mHuBERT-147: A Compact Multilingual HuBERT Model},'' \emph{arXiv preprint arXiv:2406.06371}, 2024.

\bibitem{wavlm_chen2022wavlm}
S.~Chen, C.~Wang, Z.~Chen \emph{et~al.}, ``{WavLM: Large-scale self-supervised pre-training for full stack speech processing},'' \emph{IEEE Journal of Selected Topics in Signal Processing}, vol.~16, no.~6, pp. 1505--1518, 2022.

\bibitem{encoderprompting_kashiwagi2024rapid}
Y.~Kashiwagi, H.~Futami, E.~Tsunoo, S.~Arora, and S.~Watanabe, ``{Rapid Language Adaptation for Multilingual E2E Speech Recognition Using Encoder Prompting},'' \emph{arXiv preprint arXiv:2406.12611}, 2024.

\bibitem{watanabe18_interspeech}
S.~Watanabe, T.~Hori, S.~Karita \emph{et~al.}, ``{ESPnet: End-to-End Speech Processing Toolkit},'' in \emph{Proc. Interspeech}, 2018, pp. 2207--2211.

\bibitem{voxpopuli_wang-etal-2021-voxpopuli}
\BIBentryALTinterwordspacing
C.~Wang, M.~Riviere, A.~Lee \emph{et~al.}, ``{V}ox{P}opuli: A large-scale multilingual speech corpus for representation learning, semi-supervised learning and interpretation,'' in \emph{Proceedings of the 59th Annual Meeting of the Association for Computational Linguistics and the 11th International Joint Conference on Natural Language Processing (Volume 1: Long Papers)}.\hskip 1em plus 0.5em minus 0.4em\relax Online: Association for Computational Linguistics, Aug. 2021, pp. 993--1003. [Online]. Available: \url{https://aclanthology.org/2021.acl-long.80}
\BIBentrySTDinterwordspacing

\bibitem{wenet_zhang2022wenetspeech}
B.~Zhang, H.~Lv, P.~Guo \emph{et~al.}, ``{WenetSpeech: A 10000+ hours multi-domain mandarin corpus for speech recognition},'' in \emph{ICASSP 2022-2022 IEEE International Conference on Acoustics, Speech and Signal Processing (ICASSP)}.\hskip 1em plus 0.5em minus 0.4em\relax IEEE, 2022, pp. 6182--6186.

\bibitem{peoples_galvez2021people}
D.~Galvez, G.~Diamos, J.~Ciro \emph{et~al.}, ``The people's speech: A large-scale diverse english speech recognition dataset for commercial usage,'' \emph{arXiv preprint arXiv:2111.09344}, 2021.

\bibitem{npsc_solberg2022norwegian}
P.~E. Solberg and P.~Ortiz, ``{The Norwegian parliamentary speech corpus},'' \emph{arXiv preprint arXiv:2201.10881}, 2022.

\end{thebibliography}

\end{document}